\pdfoutput=1

\documentclass[11pt]{article}

\usepackage[]{EMNLP2022}

\usepackage{times}
\usepackage{latexsym}
\usepackage[utf8]{inputenc} %
\usepackage[T1]{fontenc}    %
\usepackage{booktabs}       %
\usepackage{amsfonts}       %
\usepackage{nicefrac}       %
\usepackage{microtype}      %
\usepackage{xcolor}         %
\usepackage{graphicx}
\usepackage{subfigure}
\usepackage{amsmath}
\usepackage{multirow}
\usepackage{pifont}
\usepackage{wrapfig,lipsum,booktabs}
\usepackage{lineno}
\usepackage{colortbl}
\definecolor{Blue2}{RGB}{235, 245, 250}

\usepackage[T1]{fontenc}

\usepackage[utf8]{inputenc}

\usepackage{microtype}

\usepackage{inconsolata}

\usepackage{multirow}
\usepackage[disable]{todonotes}
\usepackage{CJKutf8}
\usepackage{booktabs}
\usepackage{subfigure}
\usepackage{graphicx}

\usepackage{enumitem}
\usepackage{amsmath}

\usepackage{amssymb}
\usepackage{graphicx}
\usepackage{scalefnt}
\usepackage{bbm}
\usepackage{float}
\usepackage{placeins} 

\usepackage[ruled,linesnumbered]{algorithm2e}

\usepackage{pgfplots}
\pgfplotsset{every tick label/.append style={font=\small}}

\usepackage{times}
\usepackage{latexsym}
\usepackage{xcolor}
\usepackage[T1]{fontenc}

\usepackage[utf8]{inputenc}

\pgfdeclareplotmark{mystar}{
\node[star,star point ratio=2.25,minimum size=6pt,
      inner sep=0pt,draw=black,solid,fill=red] {};
}

\title{CodeRetriever: Large-scale Contrastive Pre-training for Code Search}

\author{ Xiaonan Li\textsuperscript{1}\thanks{\ \ Work is done during internship at Microsoft Research Asia.}\ , Yeyun Gong\textsuperscript{2}, Yelong Shen\textsuperscript{2}, Xipeng Qiu\textsuperscript{1}\thanks{\ \ Corresponding author.}\ , \\
\textbf{Hang Zhang\textsuperscript{2}, Bolun Yao\textsuperscript{2}, Weizhen Qi\textsuperscript{2},
Daxin Jiang\textsuperscript{2}, Weizhu Chen\textsuperscript{2}, Nan Duan\textsuperscript{2} } \\
\textsuperscript{1} Shanghai Key Laboratory of Intelligent Information Processing, Fudan University \\
\textsuperscript{1} School of Computer Science, Fudan University  \textsuperscript{2}Microsoft \\
\textsuperscript{1}\{lixn20, xpqiu\}@fudan.edu.cn, \\\textsuperscript{2}\{yegong, yeshe, v-zhhang, yaobolun, weizhen djiang, wzchen, nanduan\}@microsoft.com
}

\begin{document}
\maketitle
\begin{abstract}
In this paper, we propose the CodeRetriever model, which learns the function-level code semantic representations through large-scale code-text contrastive pre-training. We adopt two contrastive learning schemes in CodeRetriever: unimodal contrastive learning and bimodal contrastive learning. For unimodal contrastive learning, we design an unsupervised learning approach to build semantic-related code pairs based on the documentation and function name. For bimodal contrastive learning, we leverage the documentation and in-line comments of code to build code-text pairs. Both contrastive objectives can fully leverage large-scale code corpus for pre-training. 
Extensive experimental results show that CodeRetriever achieves new state-of-the-art with significant improvement over existing code pre-trained models, on eleven domain/language-specific code search tasks with six programming languages in different code granularity
(function-level, snippet-level and statement-level).
These results demonstrate the effectiveness and robustness of CodeRetriever.
The codes and resources are available at \url{https://github.com/microsoft/AR2/tree/main/CodeRetriever}.
\end{abstract}

\section{Introduction}
Code search aims to retrieve  functionally relevant code given a natural language query to  boost developers' productivity ~\citep{retrieval_code_help, codesearchnet}.  
Recently, it has been shown that code pre-training techniques, such as CodeBERT ~\cite{codebert} and GraphCodeBERT ~\cite{graphcodebert}, could significantly improve code search performance via self-supervised pre-training using large-scale code corpus ~\cite{codesearchnet}.  

\begin{figure}[!t]
  \centering
    \subfigure[ Fibonacci
    ]{
    \includegraphics[width=0.45\textwidth]{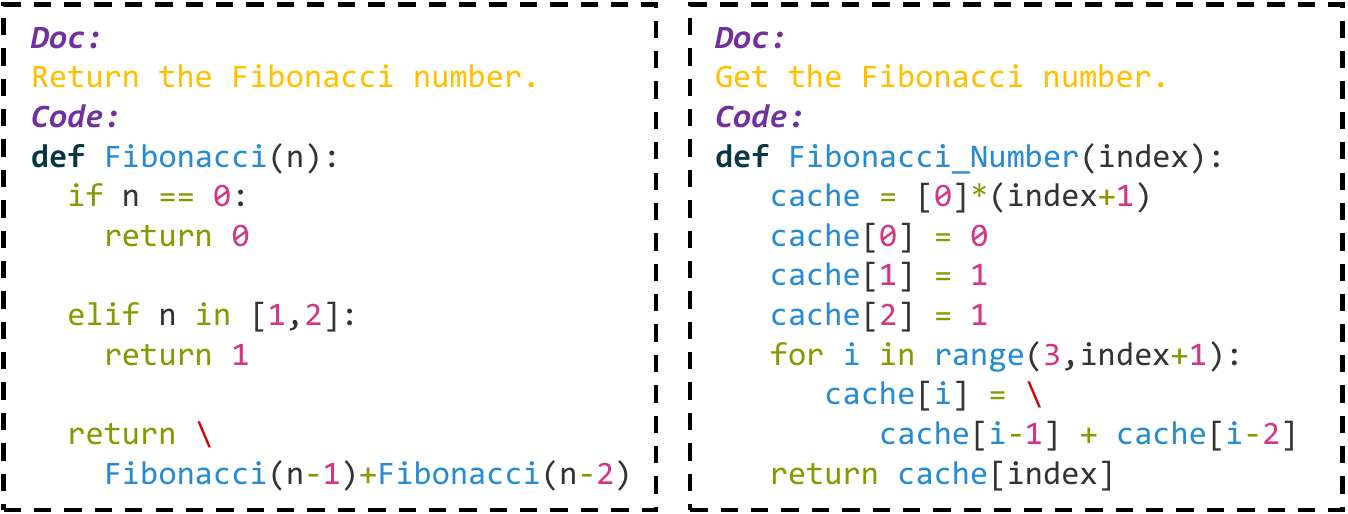}
    \label{imag_at_first_page_1}
  }
  
    \subfigure[ BubbleSort
    ]{
    \includegraphics[width=0.45\textwidth, ]{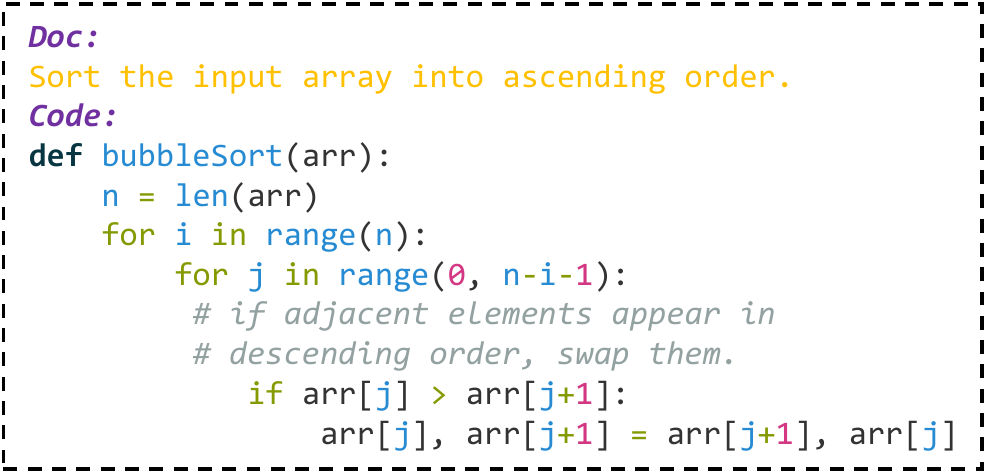}
    \label{imag_at_first_page_2}
  }
  \caption{Code examples. (a) Two different implementations of Fibonacci number algorithm; (b) Documentation, in-line comment, and code in BubbleSort implementation.}
\end{figure}

However, existing code pre-training approaches usually adopt (masked) language modeling as the training objective which targets on learning to predict (masked) tokens in a given code context ~\citep{codebert,graphcodebert,plbart,codet5}. However, this token-based approach generally results in poor code semantic representations due to two reasons. 
The first one is the anisotropy representation issue. As discussed in ~\citep{bertflow}, the token-level self-training approach causes the embeddings of high-frequency tokens clustered and dominate the representation space, which greatly limits the expressiveness of long-tailed low-frequency tokens in pre-trained models. Thus, the anisotropic representation space induces poor function-level code semantic representation ~\citep{bertflow}. In programming language, the problem of token imbalance is even more severe than that of natural language. For example, common keywords and operators such as ``='', ``\{'', and ``\}'' appear almost everywhere in Java code. 
The second one is the cross-language representation issue.  The widely used CodeSearchNet corpus \cite{codesearchnet} contains codes from six different programming languages such as Python, Java, etc.  Since the code with mixed programming languages can hardly appear within the same context, it is challenging for the pre-trained model to learn a unified semantic representation of the code with the same functionality but using different programming languages.

To address these limitations, we propose the CodeRetriever model, focusing on learning the function-level code representations, specifically for code search scenarios. The CodeRetriever model consists of a text encoder and a code encoder, which encodes text/code into separate dense vectors. The semantic relevance between code and text (or code and code) is measured by the similarity between dense vectors ~\citep{DBLP:journals/corr/abs-2004-04906, huang2013learning, shen2014a}. %

In the training of CodeRetriever, the code/text encoders are optimized by minimizing two types of contrastive losses: 
\textbf{1.Unimodal contrastive loss}, encourages the model to push codes with similar functionality closer in representation space. To estimate whether two codes are semantically close, the model needs to reason based on the given code and understand its semantics.
\textbf{2.Bimodal contrastive loss}, helps model the relevance between code and text. Since the document or comment contains rich semantic information of the code, it can encourage the model to learn better code representation from natural language.

In this work, we adopt the commonly used CodeSearchNet corpus \cite{codesearchnet} for training the CodeRetreiver. CodeSearchNet mainly contains paired dataset (a function paired with a document) and unpaired dataset (only a function). The paired dataset could be directly used for bimodal contrastive learning. For unimodal contrastive learning in CodeRetriver, we build positive code-code pairs by an unsupervised semantic-guided approach. Figure~\ref{imag_at_first_page_1} shows a code-code example: two implementations of the Fibonacci number algorithm. Moreover, the generated code-code pairs can be with different programming languages, which can mitigate the cross-language representation issues. To further take advantage of the large-scale code in unpaired data and paired data, we extract the code and in-line comment pairs to enhance the bimodal contrastive learning in CodeRetriever. Figure~\ref{imag_at_first_page_2} shows an example to indicate that the in-line comment (comment shortly) can also reflect the code's semantics and internal logic. Specifically, the underlying logic of ``if adjacent elements appear in descending order, swap them'' corresponds to sorting the input array into ascending order and such fine-grained semantic information can also help learn better code representation.

Through contrasting these unimodal and bimodal pairs, CodeRetriever can 1. learn better the function-level code semantic representation, which could alleviate the anisotropy representation issue ~\citep{simcse,consert}; 2. explicitly model the relevance of codes with different programming languages and  treat unified natural language as a fulcrum to mitigate cross-language representation issue.
We evaluate CodeRetriever on eleven code search datasets covering six programming languages, real-world scenarios and codes with different granularity (function-level, snippet-level and statement-level), and the results show that CodeRetriever achieves a new state-of-the-art performance.

\todo[inline]{Done: maybe unpaired data -> all data}

\section{Preliminary: Code Search}
CodeSearchNet corpus ~\citep{codesearchnet} is the largest publicly available code dataset. The corpus is collected from open-source non-fork GitHub repositories, which contains 2.1M paired data (a function paired with a document) and 6.4M unpaired data (only functions). 

In the literature, code-search approaches ~\citep{codesearchnet, contracode, codebert, graphcodebert} make use of the paired code-document dataset in CodeSearchNet corpus to train a siamese encoder model for language to code retrieval. However, rich unlabeled code corpus is either simply abandoned or severed as code pre-training corpus ~\citep{codebert, graphcodebert}. We argue that token-level code pre-training objectives do not explicitly learn the function-level code representation. Thus existing code pre-training models ~\citep{contracode, codebert, graphcodebert} are sub-optimal for code search.

In this work, we propose the CodeRetriever to learn the function-level code semantic representation. CodeRetriever is initialized with the code pre-trained model (i.e., GraphCodeBERT). It takes code-doc and code-comment paired data for bimodal contrastive learning, and code-code paired data for unimodal contrastive learning. After CodeRetriever's pre-training, it can serve for downstream domain/language specified datasets.
\todo[inline]{done: add coderetriever fine-tune, zero-shot usage}

\begin{figure*}[!t]
    \centering
    \includegraphics[width=0.8\textwidth, height=2in]{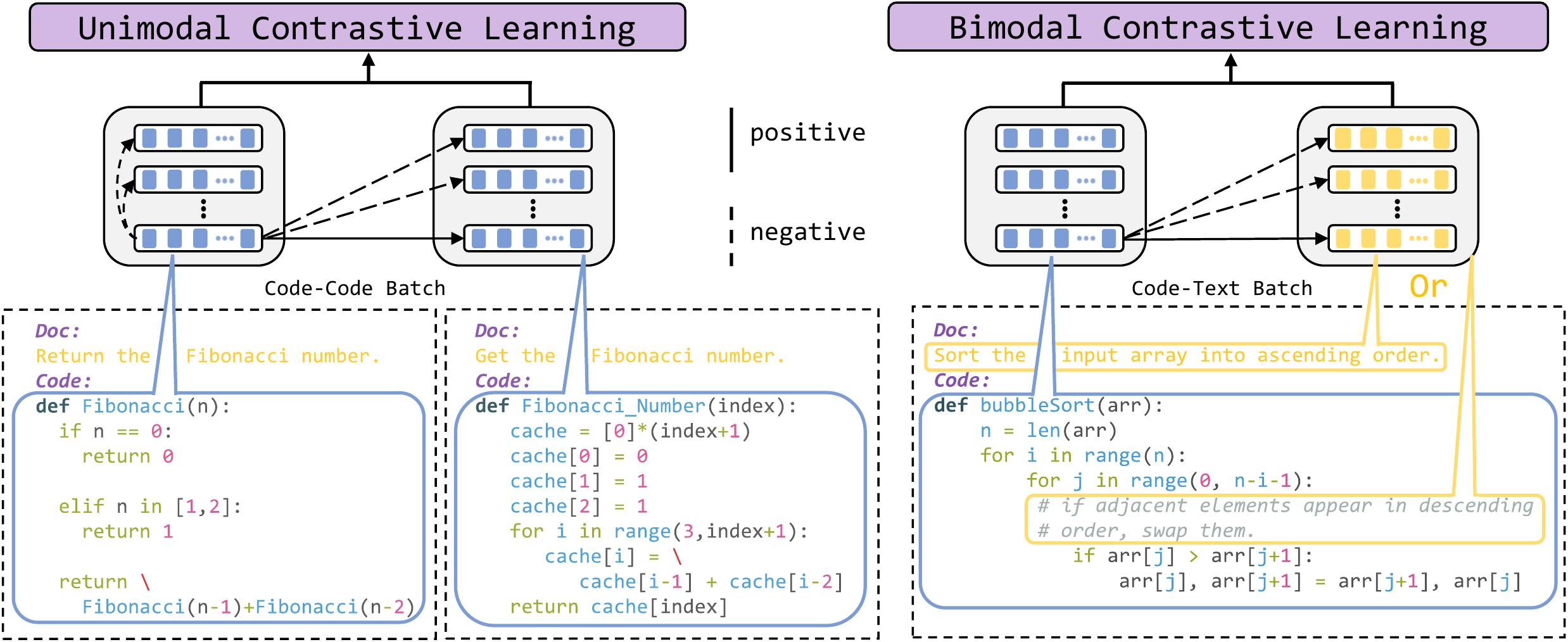}
    \caption{Unimodal and bimodal contrastive learning in CodeRetriever.}
    \label{fig:codeRetriever}
    \vspace{-10pt}
\end{figure*}

\section{Approach}
In this section, we present the model architecture and training objective of CodeRetriever. 

CodeRetriever adopts a siamese code/text encoder architecture to represent code/text as dense vectors. Let $E_{\text{code}}(\cdot; \theta)$ and $E_{\text{text}}(\cdot; \phi)$ denote code and text encoders, respectively. The semantic similarities between code-code pair $(c, c^+)$, and text-code pair $(t, c^+)$ are calculated as:
\begin{align}
    s(c, c^+) = \left\langle E_{\text{code}}(c;\theta), E_{\text{code}}(c^+;\theta) \right\rangle  \\
    s(t, c^+) = \left\langle E_{\text{text}}(t;\phi), E_{\text{code}}(c^+;\theta) \right\rangle ,
\end{align}
where $\left\langle , \right\rangle$ indicates cosine similarity operation.

\subsection{Unimodal Contrastive Learning}
Given a paired code-code training sample $(c, c^+)$, the unimodal contrastive loss is given by:
\begin{align}
    \mathcal{L}_{\mathrm{uni}} = -\operatorname{ln}\frac{\operatorname{exp}\left(\tau s(c, c^+)\right) }{ \sum_{c' \in \mathbb{C}} \operatorname{exp}\left(\tau s(c, c')\right)  } \quad,
    \label{eq:uni}
\end{align}
where $\tau$ is the temperature, for simplicity, we let $\tau = 1$; set $\mathbb{C}$ consists of the paired code $c^+$ and $N-1$ unpaired code samples obtained by in-batch negative sampling \cite{DBLP:journals/corr/abs-2004-04906}. In particular, one batch can consist of hybrid programming languages, which can help the pre-trained model to learn a unified semantic space of codes with different programming languages.

\subsection{Bimodal Contrastive Learning}
Given a paired text-code training instance $(t, c^+)$, the bimodal contrastive loss is defined as the same manner:
\begin{align}
    \mathcal{L}_{\mathrm{bi}} = -\operatorname{ln}\frac{\operatorname{exp}\left(\tau s(t, c^+)\right) }{ \sum_{c' \in \mathbb{C}} \operatorname{exp}\left(\tau s(t, c')\right)  } \quad,
\end{align}

where the definitions of $\tau$ and $\mathbb{C}$ are the same as in eqn. ~\ref{eq:uni}. The codes of the text-code batch also consist of hybrid programming languages, which can help align the semantic space of different programming languages and natural language. Since the document or comment reflects the functionality and crucial semantic information of source code, such positive pairs can help model better understand the semantics of code.

\subsection{Overall Pre-training Objective}
As illustrated in Figure ~\ref{fig:codeRetriever}, CodeRetreiver takes two types of text-to-code for bimodal contrastive training, which are code-document and code-comment. Therefore, we use $\mathcal{L}_{\mathrm{bi}}^1$ and $\mathcal{L}_{\mathrm{bi}}^2$ to denote code-document and code-comment contrastive loss, respectively.
The overall pre-training objective for CodeRetreiver is:

\begin{align}
    \mathcal{L}(\theta, \phi) =  \mathcal{L}_{\mathrm{uni}} +  \mathcal{L}^1_{\mathrm{bi}} +  \mathcal{L}^2_{\mathrm{bi}}
\end{align}

\section{Building Positive Pairs}

\begin{figure*}[!t]
  \centering
    \subfigure[]{
    \includegraphics[width=0.7\textwidth]{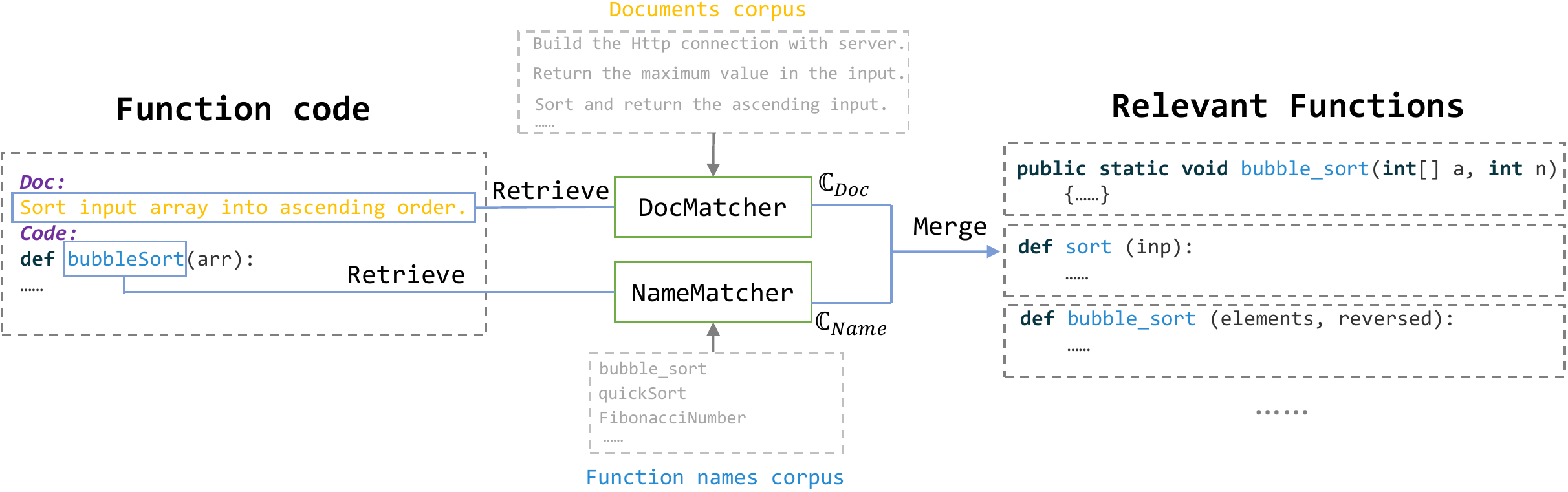}
    \label{data_pipeline_1}
  }
  \unskip\ \vrule
    \subfigure[]{
    \raisebox{0.6\height}
    {
    \includegraphics[width=0.2\textwidth]{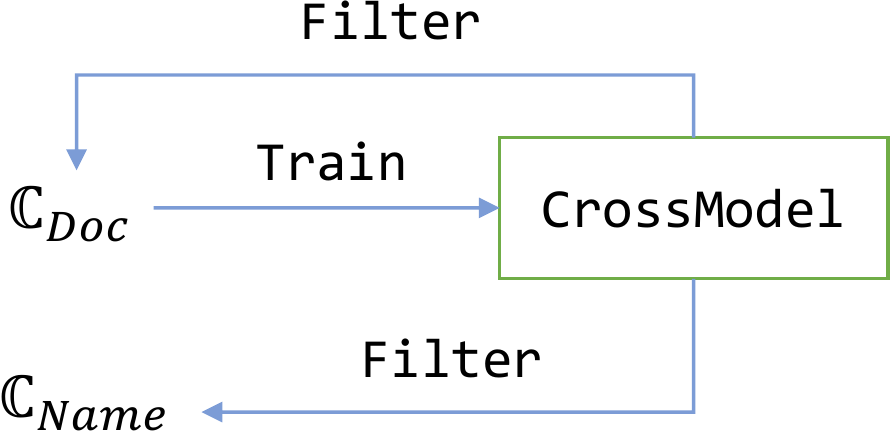}
    }
    \label{data_pipeline_2}
  }
  \vspace{-10pt}
  \caption{The illustration of building code-code pairs. (a) Step 1. Collect noisy code-code pairs through function name match and documentation match; (b) Step 2. Denoise code-code pairs with CrossModel.}
  \label{data_pipeline}
  \vspace{-10pt}
\end{figure*}

\subsection{Code-Document}
Documents of source codes usually can provide rich semantic information and highly describe the functionality of codes. For example, in Figure~\ref{imag_at_first_page_2}, the document “Sort the input array into ascending order.” clearly summarizes the goal of the code, which can help the model to better understand the code. So we take code $c$ and its corresponding document $t$ as positive pairs. Thus we can not only help model better understand code but also align different programming languages' representation through the unified natural language description as a pivot.
\subsection{Code-Comment}
Unlike documents, the in-line comments widely exist in unpaired code. As shown in Figure~\ref{imag_at_first_page_2}, it can reflect the code's internal logic and contains fine-grained semantic information, despite certain noisy signals. So we consider code-comment as positive pairs to further help model to learn better code representation. 
In this section, we introduce how we build code-comment pairs.
We first leverage the code parser (tree-sitter) to split the code-block into two parts: pure code and the corresponding in-line comments. Then we perform post-processing as follows to filter noisy paired samples to obtain the code-comment corpus:
\begin{itemize}[noitemsep,topsep=0pt,parsep=2pt,partopsep=0pt,
leftmargin=0.4cm
]
    \item We merge comments with continuous lines into one comment. This is inspired by the phenomenon where developers usually write a complete comment into multiple-lines to make it easier to read, like in Figure~\ref{imag_at_first_page_2}.
    \item Comments with little information are removed, including: 1) shorter than four tokens; 2) comments beginning with ``TODO''; 3) comments for automated code checking, like ``Linter~$\cdots$''\footnote{Linter is a static analysis tool for checking code.}. 4) non-text comments, i.e., commented code. 
    \item Functions with little semantic information are removed such as functions with names ``\_\_getter\_\_'', ``\_\_setter\_\_'' etc, are removed. 
\end{itemize}
After cleaning, we collect about 1.9 million code-comment pairs. The detailed statistics of the overall code-text corpus can be seen in Appendix~\ref{sec:appendix_pretrain_dataset}.

\subsection{Code-Code}
Code-code paired datasets can provide explicit training signals for models to learn the semantic representation of code. However,
it is challenging to build large-scale and high-quality semantically relevant code-to-code pairs from an unlabeled corpus. To a specific functionality, there are a lot of ways to implement it and the resulting code can be full of diversity. They can have totally different logic, libraries invoked, and identifier names. Even for experienced developers, it's challenging and time-consuming for them to assess the semantic similarity of two codes, which makes human annotation costly and not scalable. Although two codes of the same functionality can have different implementations, their documentations or function names can be very similar, as shown in Figure~\ref{imag_at_first_page_1}. Inspired by this phenomenon, we propose the unsupervised techniques as following to collect a large-scale code-to-code corpus.

\begin{table*}[t]
\small
\centering
\setlength{\tabcolsep}{3.1mm}

\begin{tabular}{@{}lccccccc@{}}
\toprule
\textbf{Lang}    & \textbf{Ruby}       & \textbf{Javascript} & \textbf{Go}     & \textbf{Python} & \textbf{Java}   & \textbf{PHP} &\textbf{Overall}   \\ \midrule

ContraCode~\citep{contracode}       & -                   & 30.6                & -               & -               & -               & -              & - \\
SyncoBERT~\citep{syncobert}        & 72.2                & 67.7                & 91.3            & 72.4            & 72.3            & 67.8           &74.0 \\
CodeBERT~\citep{codebert}         & 67.9                & 62.0                & 88.2            & 67.2            & 67.6            & 62.8           &69.3 \\
GraphCodeBERT~\citep{graphcodebert} \hspace{-5pt}   & 70.3                & 64.4                & 89.7            & 69.2            & 69.1            & 64.9           &71.3 \\
UniXcoder~\citep{unixcoder} &74.0&68.4&91.5&72.0&72.6&67.6&74.4 \\

CodeRetriever (In-Batch Negative)  &  75.3                &  69.5                &  91.6             &  73.3            &  74.0            &  68.2          &  75.3   \\
CodeRetriever (Hard Negative)  &  75.1                &  69.8                 &  92.3            &  74.0            &  74.9            &  69.1         &  75.9   \\
CodeRetriever (AR2) &  \textbf{77.1}               &  \textbf{71.9}               &  \textbf{92.4}           &  \textbf{75.8}           &  \textbf{76.5}        &     \textbf{70.8}  &  \textbf{77.4}\\ 
\bottomrule
\end{tabular}
\caption{The comparison on the CodeSearch dataset. We get the ContraCode's result by fine-tuning the released checkpoint~\citep{contracode}. Other results of compared models are reported by previous papers.}
\label{main_result_1}
\end{table*}

\begin{table*}
\small
\centering
\setlength{\tabcolsep}{3.4mm}
\begin{tabular}{@{}lcccccc@{}}
\toprule
            
\textbf{Dataset}  & \textbf{Adv}        & \textbf{CoSQA}  & {\textbf{CoNaLa}} & \textbf{SO-DS}  & \textbf{StaQC} &\textbf{Overall} \\ \midrule

SyncoBERT~\citep{syncobert}                      & 38.1                & -               & -               & -               & -         & -      \\
CodeBERT~\citep{codebert} & 27.2                & 64.7            & 20.9            & 23.1            & 23.4            & 31.9\\
GraphCodeBERT~\citep{graphcodebert} & 35.2                & 67.5            & 23.5            & 25.3            & 23.8           &35.1 \\
UniXcoder~\citep{unixcoder} & 41.3&70.1 &- & -& -& -\\
CodeRetriever (In-Batch Negative) &  43.0  &  70.6  &  29.6       &  27.1 &  \textbf{25.5}  &  39.0 \\
CodeRetriever (Hard Negative) &  45.1  &  74.1  &  \textbf{29.9} &  31.8 &  24.6 &  41.1 \\
CodeRetriever (AR2)  &\textbf{46.9 }                &  \textbf{75.4} &  29.1 &  \textbf{33.9} &  24.2 &  \textbf{41.9}\\ 
\bottomrule

\end{tabular}
\caption{The comparison on datasets that are closer to the real scenario. The results of Compared models on the Adv dataset and UniXcoder on CosQA are reported by previous papers, other results are from our implementation since they are not reported previously.}
\label{main_result_2}
\vspace{-10pt}
\end{table*}

\paragraph{Step 1. Collect noisy code-code pairs by maching function name and documentation.}\hspace{25pt}
\textbf{1)} We adopt the recently proposed unsupervised method, SimCSE~\citep{simcse}, to train with the function name corpus, obtain ``NameMatcher'' model; and train with documentation corpus to obtain ``DocMatcher'' model; Both ``NameMatcher'' and ``DocMatcher'' are dense retrieval models. For example, given a function name, ``NameMatcher'' could be able to retrieve top-K relevant function names in the corpus. We refer readers to its original paper~\citep{simcse} for more details.
\textbf{2)} For any given function in the corpus,  we retrieve its relevant functions through function name matching using the ``NameMatcher''. The similar manner is applied to ``DocMatcher'', which collects code-code pairs by matching their corresponding documentations. We denote the code-code pairs collected through ``DocMatcher'' as $\mathbb{C}_{\text{Doc}}$, and use $\mathbb{C}_{\text{Name}}$ to indicate the code-code pairs collected through ``NameMatcher''. We only keep code-code pairs if their retrieval scores (by ``NameMatcher'' and ``DocMatcher'') are greater than threshold $(0.75)$. 
\paragraph{Step 2. Denoise Code-code pairs with CrossModel.} 
The code-code sets $\mathbb{C}_{\text{Name}}$ and $\mathbb{C}_{\text{Doc}}$ collected from Step 1 can be noisy, especially for $\mathbb{C}_{\text{Name}}$ as functions with the same function name can have different functionalities. In this step, we train a binary classifier model, CrossModel ($\text{M}_c$), for filtering noisy code-code pairs. 
\textbf{1)} We take the code-code pairs $\mathbb{C}_{\text{Doc}}$, which is less noisy, as the training set to train the CrossModel $\text{M}_c$. 
It takes the concatenation of code-code pair as input and is more powerful for predicting their relevant score (range from 0 to 1) via deep token interaction. In the training of $\text{M}_c$,  we use set $\mathbb{C}_{\text{Doc}}$ as positive training instances while sampling random code-code pairs as negative instances.  
\textbf{2)} We remove the code-code pairs in $\mathbb{C}_{\text{Doc}}$ and $\mathbb{C}_{\text{Name}}$ if their prediction scores by $\text{M}_c$ are smaller than certain threshold. Let $\mathbb{C}_{\text{Name}}^*$ and $\mathbb{C}_{\text{Doc}}^*$ be the denoised subsets of $\mathbb{C}_{\text{Name}}$ and $  \mathbb{C}_{\text{Doc}}$. The final code-code corpus is the joint of set  $\mathbb{C}_{\text{Name}}^*$ and $\mathbb{C}_{\text{Doc}}^*$. Since we 
take the natural language as the anchor to get $\mathbb{C}_{\text{Name}}^*$ and $\mathbb{C}_{\text{Doc}}^*$, the code pair can have different programming languages and mitigate the cross-language representation issue.

\todo[inline]{maybe description about using only c\_doc can reduce the noise of M\_c can be polished.}

We show the process of Step 1 and Step 2 in 
Figure~\ref{data_pipeline_1} and Figure~\ref{data_pipeline_2}, respectively. Overall, the collected code-code corpus contains 23.4 million pairs. We provide a more detailed description on building code-code corpus, involved hyper-parameters and detailed cross-language statistics of code-code pairs in Appendix~\ref{sec:appendix_code_code_pairs_algorithm}, \ref{sec:appendix_building_code_code_hyper} and \ref{sec:appendix_unimodal_pairs}.

\section{Experiment}

For fair comparison, CodeRetriever adopts the same model architecture as previous works~\citep{codebert,graphcodebert}. CodeRetriever shares parameters of code encoder and text encoder. It contains 12 layers Transformer with hidden size of 768 and attention heads of 12. To accelerate the training process, we initilaize CodeRetriever with the released parameters of GraphCodeBERT~\citep{graphcodebert}. We show more details in Appendix~\ref{sec:appendix_implementation_details}.

\subsection{Benchmark Datasets}

We evaluate CodeRetriever on several code search benchmarks, including \textbf{CodeSearch}~\citep{codesearchnet,graphcodebert}, \textbf{Adv}~\citep{codexglue}, \textbf{CoSQA}~\citep{cosqa}, \textbf{CoNaLa}~\citep{conala}, \textbf{SO-DS}~\citep{sods}, \textbf{StaQC}~\citep{staqc}. The CodeSearch benchmark contains six datasets with different programming languages. %
The Adv dataset normalizes the method names and variable names in the dev/test set, which makes it more challenging. CoNaLa, SO-DS, and StaQC are collected from stackoverflow questions, and CoSQA are collected from web search engines. Therefore, the queries in CoSQA, CoNaLa, SO-DS, and StaQC are closer to the real code-search scenario compared with Adv and CodeSearch. Meanwhile, CoNALA, SO-DS and StaQC contain the code with different granularity, i.e., statement-level and snippet-level. The statistics of these benchmark datasets are listed in Appendix~\ref{sec:appendix_finetune_dataset}. Following previous works~\citep{codebert,graphcodebert}, we use Mean Reciprocal Rank (MRR)~\citep{mrr1} as the evaluation metric on all benchmark datasets.

\subsection{Experiment: Fine-Tuning}

In the fine-tuning experiments, CodeRetriever and other code pre-trained models are fine-tuned on the eleven language/domain-specific code search tasks, each task provides a set of labeled query-code pairs for model adaptation. 
\footnote{For the CodeSearch benchmark, although it has overlapping with the paired data of pre-training corpus, fine-tuning on it is different from CodeRetriever's bimodal contrastive learning. In detail, fine-tuning on CodeSearch only covers one specific programming language's query-code pair while CodeRetriever's bimodal contrastive learning covers 1. six hybrid programming languages for unifying their semantic space 2. extra comment-code pair for further taking advantage of the unpaired data.
}

\subsubsection{Fine-tuning}
Previous works on dense text retrieval ~\citep{dpr,ance,rocketqa} show that the strategy of selecting negative samples could greatly affect the model performance in contrastive learning tasks. Therefore, we explore the following three approaches for CodeRetriever fine-tuning:
\textbf{1.In-Batch Negative.}
 For a <query, code> pair in a batch, it uses other codes in the batch as negatives ~\citep{dpr}. Existing code pre-trained models take in-batch negative as the default fine-tuning method ~\cite{codebert,graphcodebert,syncobert}. 
\textbf{2.Hard Negative.} It can pick ``hard'' representative negative samples other than random negatives. Compared with in-batch negative, the hard negative training is more efficiency~\citep{dpr}, which is widely used in text dense retrieval. We follow ~\citet{condenser} for hard negative fine-tuning.
\textbf{3.AR2.} 
It is a recently proposed training framework for dense retrieval~\citep{ar2}. It adopts an adversarial-training approach to select ``hard'' negative samples iteratively. In this paper, we focus on using AR2 to enhance the siamese encoder for code search.

In fine-tuning experiments, we conduct grid search over learning-rate in \{2e-5, 1e-5\}, batch-size in \{32, 64, 128\}. Training epoch, warm-up step, and weight decay are set to 12, 1000, and 0.01, respectively on all tasks. We report the average results under 3 different random seeds.
The hyper-parameters for AR2 training are listed in Appendix~\ref{sec:appendix_ar2_hyper}.

We compare CodeRetriever with state-of-the-art pre-trained models, including:
\mbox{\textbf{CodeBERT}}~\citep{codebert}, pre-trained with MLM and replaced token detection tasks; 
\textbf{GraphCodeBERT}~\citep{graphcodebert}, which integrates data flow baesed on CodeBERT. \textbf{SynCoBERT}~\citep{syncobert}, pre-trained on code-AST pairs with contrastive learning; \textbf{ContraCode}~\citep{contracode}, pre-trained with contrastive learning through semantic-preservin code transformation on Javascript corpus. \textbf{UniXcoder}~\citep{unixcoder} is adpated from UniLM and pre-trained on unified cross-modal data like code, AST and text.

\subsubsection{Results}
Table~\ref{main_result_1} and Table ~\ref{main_result_2} show the performance comparison on all benchmark datasets. First, we report the performance of CodeRetriever (In-Batch Negative), which uses the same finetuning approach as other baselines to ensure a fair comparison. It shows that CodeRetriever obtains the best overall performance compared with all other compared approaches.
Specifically, CodeRetriever improves over GraphCodeBERT by 4.0 average absolute points on the CodeSearch dataset, which demonstrates the effectiveness of CodeRetriever. Meanwhile, CodeRetriever outperforms the previous state-of-the-art model, UniXcoder~\citep{unixcoder}, on all tasks with reported results. On the Adv, CoSQA, CoNaLa, SO-DS and StaQC datasets, CodeRetriever also outperforms baselines models, which shows that CodeRetriever consistently outperforms baseline models in various scenarios.
\todo[inline]{Done: add description 1.real scenario 2.different granularity}

Comparing different fine-tuning approaches, we can see that the AR2 is generally better than In-Batch Negatives and Hard Negatives. i.e., CodeRetriever(AR2) improves over In-Batch Negative by 3.0 absolute points in average, and improves over Hard Negative by 1.1 absolute points in average. The experiment results suggest that selecting a good fine-tuning approach is also very important for downstream code search tasks. From Table ~\ref{main_result_2}, an interesting observation is that In-Batch Negative outperforms Hard Negatives and AR2 on StaQC benchmark. A possible explanation is StaQC contains more false query-code pairs in the training set compared with other benchmarks, as it is collected from stackoverflow through a rule-based method without any human annotations, and In-Batch Negative is more noise-tolerant than AR2 and Hard-Negative.

\subsection{Analysis}
\subsubsection{Low-Resource Code Search}
We evaluate the performance of CodeRetriever on low resource scenario, i.e., only a few hundreds of paired query-code data for fine-tuning. Table~\ref{few_shot_result} shows the results of CodeRetriever and GraphCodeBERT in the low-resource setting on CoSQA dataset, where the number of training examples is varied from 500 to FULL (19K). We can see that CodeRetriever could reach more reasonable performance in low-resource setting than GraphCodeBERT.

\subsubsection{Cross-Language Code Search}
\paragraph{Performance} Since building pairs of real user query and code is labor-intensive and costly, Existing code search datasets of real-world scenario only cover few programming language, including Python~\citep{staqc,sods,conala,cosqa}, Java~\citep{codesearchdataset_java_1,codesearchdataset_java_2} and SQL~\citep{staqc}. Here, we introduce a new setting, cross-language code search, where we fine-tune model with `A' programming language and test it on `B' programming language. This can alleviate the data scarcity problem of other programming languages. For evaluating our method on this setting, we finetune the model with query-Python corpus (CoNaLa~\cite{conala}) and evaluate it with query-Java test set ~\citep{codesearchdataset_java_2}. The queries in the Python corpus and Java corpus are both collected from stackoverflow. In Table~\ref{cross_result}, it shows that unimodal contrastive loss in CodeRetriever significantly helps the cross-language code search task. By combining bimodal contrastive loss, CodeRetriever could obtain better performance. This result indicates CodeRetriever's potential utility for real scenarios.

\begin{figure}[!t]
  \centering
    \subfigure[without $\mathcal{L}_{\mathrm{uni}}$]{
    \includegraphics[width=0.2\textwidth]{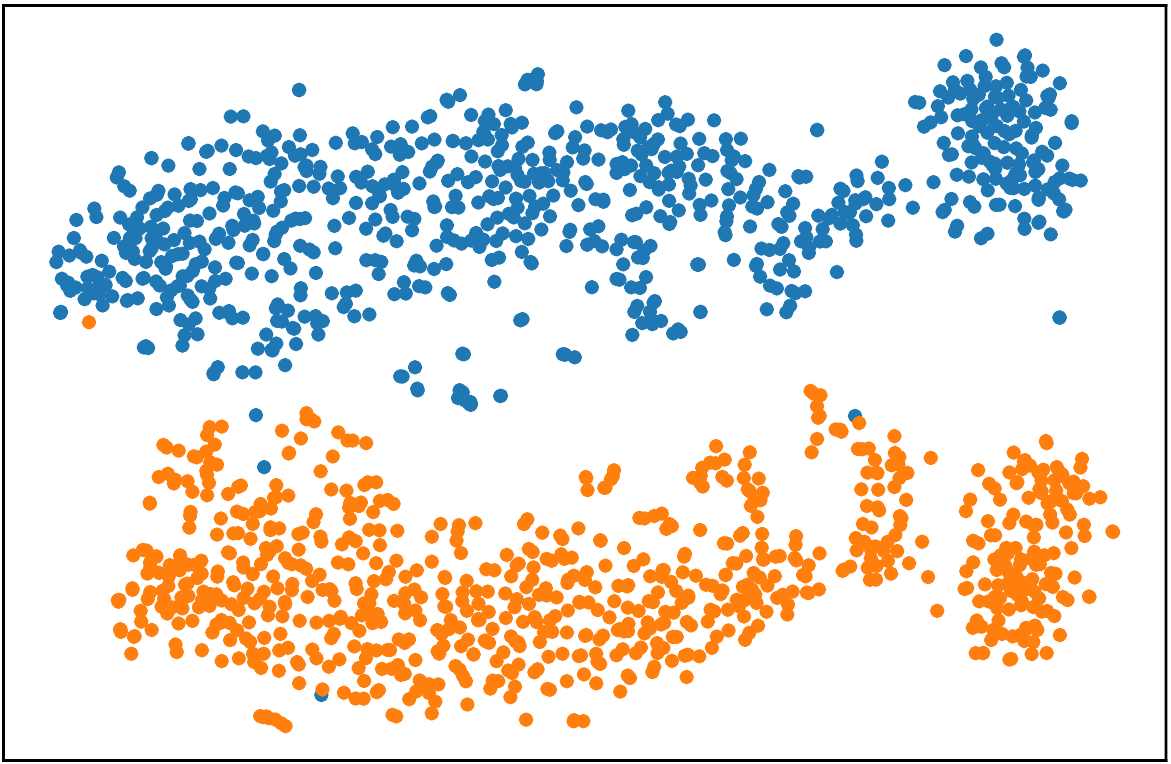}
    \label{cross_1}
  }
    \subfigure[with $\mathcal{L}_{\mathrm{uni}}$]{
    \includegraphics[width=0.2\textwidth]{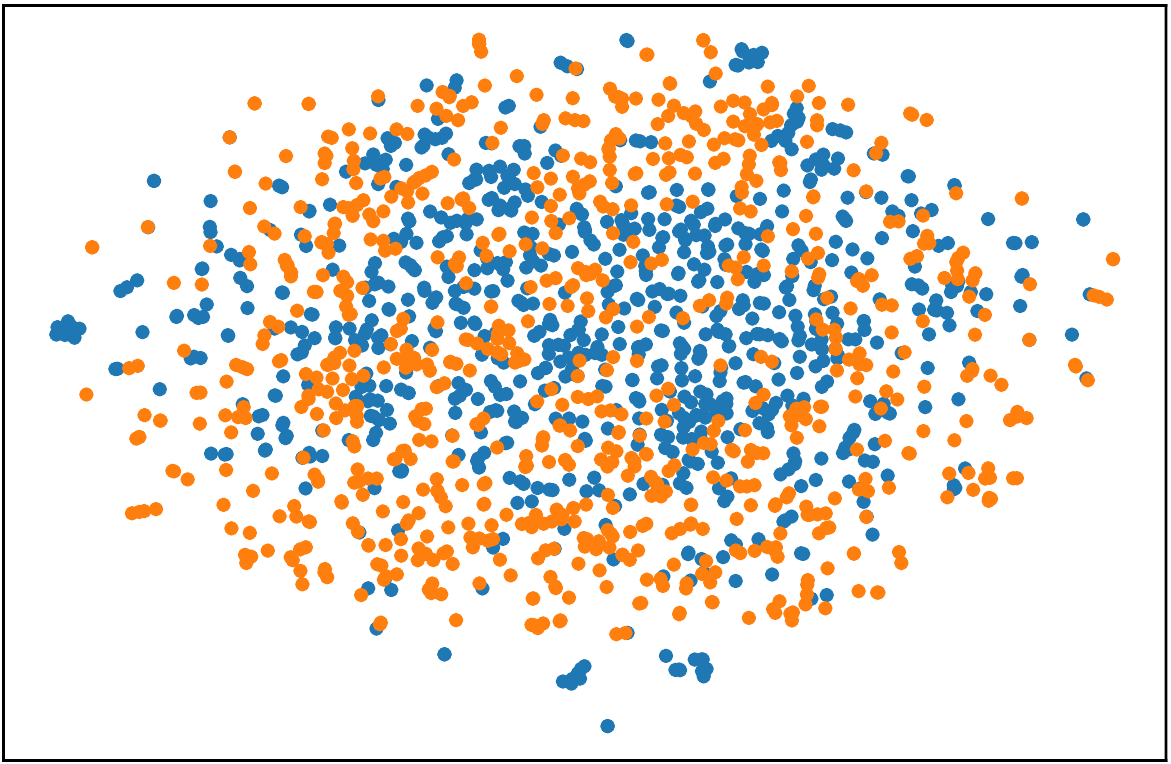}
    \label{cross_2}
  }
  \caption{The 2-D visualizations of Python and Java's representation, where \includegraphics[width=.1cm]{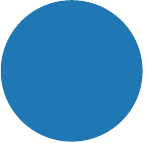} and \includegraphics[width=.1cm]{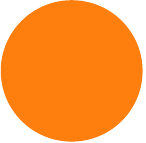} represent samples of Java and Python, respectively. }
  \label{cross_tsne}
\end{figure}

\paragraph{Visualization} 
To further analyze the effect of unimodal contrastive learning, we visualize the 2-D latent space of representations with or without unimodal contrastive learning by t-SNE~\citep{tsne}. 
In the Figure~\ref{cross_1}, we can see the representations of Java and Python code  appear in two separate clusters for the model without unimodal contrastive learning (GraphCodeBERT) while in Figure~\ref{cross_2}, their representation space are overlapped. It shows that the unimodal contrastive learning helps to learn a unified representation space of code with different programming languages.

\begin{table}[t]
\small
\centering

\setlength{\tabcolsep}{10mm}

\begin{tabular}{@{}ll@{}}
\toprule
Method        & MRR  \\ \midrule
GraphCodeBERT & 41.6 \\
CodeRetriever ({\scriptsize $\mathcal{L}_{\mathrm{uni}}$}) & 48.4 \\
CodeRetriever ({\scriptsize $\mathcal{L}_{\mathrm{uni}} +\mathcal{L}_{\mathrm{bi}}$}) & 53.3 \\
\bottomrule
\end{tabular}
\caption{The comparison on cross language code search.}

\label{cross_result}
\end{table}

\begin{table}[]

 \small
\begin{tabular}{@{}lccccl@{}}
\toprule
Train Size    & 500  & 1000 & 2000 & 4000 & FULL \\ \midrule
GraphCodeBERT & 43.2 & 49.9 & 54.0   & 57.2 & 67.5 \\
CodeRetriever   & 54.7 & 55.6 & 58.4 & 60.5 & 70.6 \\ \bottomrule
\end{tabular}
\caption{The performance comparison on CosQA with different training size.}
\label{few_shot_result}
\vspace{-15pt}
\end{table}

 \begin{figure}[t]
 \centering

       \begin{subfigure}
  \centering
  \begin{tikzpicture}[scale=0.45]
  \centering
    \begin{axis}[
    legend cell align={left},
    yscale=0.55,
      xlabel={$l_{\mathrm{uniform}}$ / Training Steps $(k)$},
      grid=major,
      legend style={at={(0.95,1.4)}},
    ]

    \addplot [color=red,mark=square*]coordinates { 
        (0,-0.5347)
    };
    \addlegendentry{{\small GraphCodeBERT}}
    
    \addplot [color=teal,mark=square]coordinates { 
        (0,-0.5347)
        (10,-1.44)
        (20,-1.74)
        (30,-1.93)
        (40,-2.01)
        (50,-2.07)
        (60,-2.13)
        (70,-2.18)
        (80,-2.22)
        (90,-2.25)
        (100,-2.2714)
    };
    \addlegendentry{{\small CodeRetriever}}

    \end{axis}
  \end{tikzpicture}
\end{subfigure}
\begin{subfigure}
  \centering
  \begin{tikzpicture}[scale=0.45]
  \centering
    \begin{axis}[
    legend cell align={left},
    yscale=0.55,
      xlabel={$l_{\mathrm{align}}$ / Training Steps $(k)$},
      grid=major,
    ]

    \addplot [color=red,mark=square*]coordinates { %
        (0,1.2279)
    };
    \addlegendentry{{\small GraphCodeBERT}}
    
    \addplot [color=teal,mark=square]coordinates { %
        (0,1.2279)
        (10,1.052)
        (20,0.9832)
        (30,0.952)
        (40,0.941)
        (50,0.93)
        (60,0.92)
        (70,0.91)
        (80,0.907)
        (90,0.903)
        (100,0.9013)
    };
    \addlegendentry{{\small CodeRetriever}}

    \end{axis}
  \end{tikzpicture}
\end{subfigure}

  \vspace{-10pt}  
  \caption{The alignment and uniformity curve.}

  \label{align_and_uniform}
  \vspace{-5pt}
 \end{figure}
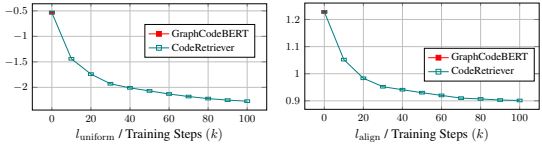

\subsubsection{Code-to-Code Search Results}
We fine-tune and evaluate CodeRetriever on code-to-code search task. In this task, given a code, the model is asked to return a semantically related code. We conduct experiment on POJ-104 dataset~\citep{poj104,codexglue} and use the same hyper-parameters as previous works \citep{codexglue}. We evaluate by Mean Average Precision (MAP), as shown in table~\ref{code_to_code_finetuning}.
We see that CodeRetriever outperforms other pre-trained models, which demonstrates its scalability and potentiality for other code understanding tasks.

\begin{table}[]
\small
\centering
\begin{tabular}{lc}
\toprule
Model         & \multicolumn{1}{l}{MAP@R} \\ \midrule
RoBERTa~\citep{roberta}       & 76.67                     \\
CodeBERT~\citep{codebert}      & 82.67                     \\
GraphCodeBERT~\citep{graphcodebert} & 85.16                     \\
SynCoBERT~\citep{syncobert}     & 88.24                     \\
DISCO~\citep{boost_contra_code}         & 82.77                     \\
Corder~\citep{corder}        & 84.10                     \\
CodeRetriever   & \textbf{88.85}                     \\ \bottomrule
\end{tabular}
\caption{The performance comparison on the code-to-code retrieval task~\citep{poj104}. Compared models' results are from previous papers~\citep{syncobert,boost_contra_code,corder}.}
\label{code_to_code_finetuning}
\end{table}

\subsubsection{Uniformity and Alignment}
To study the effect of CodeRetriever on the function-level representation space, we use the alignment and uniformity metrics~\citep{uniform_and_align_metric} to see function-level representation distribution changes during training, shown in Figure~\ref{align_and_uniform}. We see that the uniformity loss of CodeRtriever descends gradually, indicating the anisotropy is alleviated. We find that the alignment loss also has a declining trend, which shows the training of CodeRetriever can help align the representation of code and natural language and better understand them. The two metrics indicate that the CodeRetriever reduces the gap between pre-training and fine-tuning, compared with previous code pre-trained models.

\subsubsection{Ablation Study}
To understand the effect of each component in CodeRetriever, we conduct ablation study on the CodeSearch Java dataset and SO-DS. We start from the initial model and add components of CodeRetriever to it one-by-one.
We find that using code-code pairs without denoising for unimodal contrastive learning brings slight performance degradation while with denoising, it achieves significant performance improvement. This demonstrates the effectiveness of the denoising step and shows that the unimodal contrastive learning depends on the quality of positive pairs construction. Here, we verify a simple and effective positive pairs construction method, we leave the development of more powerful method as future work. From the results of using doc-code and comment-code for bimodal contrastive learning, we see that the model achieves further performance improvement, which shows the bimodal contrastive learning can 
leverage crucial semantic information in documents or comments to help better understand the code.
\begin{table}[]
 \small
    \centering
    
\begin{tabular}{@{}lcc@{}}
\toprule
Methods                       & \multicolumn{1}{l}{CodeSearch} & \multicolumn{1}{l}{SO-DS} \\ \midrule
GraphCodeBERT (Our Initial)                   & 69.1                           & 25.3                      \\
+ Code-to-Code (no denoising) & 68.9                           & 25.2                      \\
+ Code-to-Code (denoising)    & 71.1                           & 25.9                      \\
\ \ \ \ + Doc-to-Code                 & 72.2                           & 26.6                      \\
\ \ \ \ \ \ \ \ + Comment-to-Code             & 74.0                           & 27.1                      \\ \bottomrule
\end{tabular}
\caption{Ablation study.}

    \label{ablation}
    \vspace{-20pt}
\end{table}

\vspace{-2pt}
\section{Related Work}
\vspace{-3pt}

\paragraph{Token-Level Code Pre-training}
Token-level pre-trained models have been widely-used for the programming languages. 
\citet{scelmo} pre-train ELMo on JavaScript corpus for program-repair task. \citet{cubert} use a large-scale Python corpus to pre-train the BERT model. C-BERT~\cite{cbert} is pre-trained on a lot of repositories in C language and achieves significant improvement in abstract syntax tree (AST) tagging task. CodeBERT~\cite{codebert} is pre-trained by the masked language model and replaced token detection tasks on the text-code pairs of six programming languages. GraphCodeBERT~\cite{graphcodebert} introduces the information of dataflow based on CodeBERT. Besides these BERT-like models, CodeGPT~\cite{codegpt}, PLBART~\cite{plbart}, CoTexT~\cite{cotext} and CodeT5~\cite{codet5} are pre-trained for code generation tasks based on the GPT, BART~\citep{bart} and T5~\citep{nl_t5} respectively. However, token-level objectives cause the anisotropy problem~\citep{unixcoder} and have a gap with code search which is based on function-level representations. Different from these works, CodeRetriever utilizes the contrastive-learning framework to enhance the function-level representation.
\vspace{-5pt}
\paragraph{Contrastive Learning for Code} 
Recently, several works try to use contrastive learning on programming language, whose key is building effective positive or negative samples. ContraCode~\cite{contracode} and Corder~\cite{corder} use semantics-preserving transformations, such as identifier renaming and dead code insertion to build positive pairs.~\citet{boost_contra_code} develop bug-injection to build hard negative pairs. 
SynCoBERT~\cite{syncobert} and Code-MVP~\citep{code_mvp} build positive pairs through programs'compilation process like AST and CFG. However, their methods usually generate positive samples with similar structure or the same variable names as the original code, whose naturalness  and diversity is limited by hand-written rules\cite{scoder}. In CodeRetriever, we construct positive pairs from code-code, code-documentation, and code-comment. For code-code, we design a more natural and diverse positive pairs construction method based on real-world codes. 
\label {related_work_code_contrastive}
\vspace{-3pt}
\section{Conclusion}
\vspace{-3pt}
In this paper, we introduce CodeRetriever that combines unimodal and bimodal contrastive learning as pre-training tasks for code search. For unimodal contrastive learning, we propose a semantic-guided method to build positive code pairs. For bimodal contrastive learning, we utilize the document and in-line comment to build positive text-code pairs. Extensive experimental results on several publicly available benchmarks show that the proposed CodeRetriever brings significant improvement and achieves new state-of-the-art on all benchmarks. Further analysis results show that CodeRetriever is also powerful on low resource and cross-language code search tasks, and demonstrate the effectiveness of unimodal and bimodal contrastive learning.

\section*{Limitations}
CodeRetriever mainly has two limitations:

1) Due to the limited computing infrastructure, only GraphCodeBERT is used as the initialization model in the experiments. We leave experiments based on other code pre-trained models such as UniXcoder~\citep{unixcoder} as future work.

2) The code-code pairs and code-comment pairs still contain certain noise. We will explore stronger denoising methods in future work.

3) We pre-train CodeRetriever on the CodeSearchNet corpus. In future work, we will consider using more pre-training corpora such as full Github repositories.

\bibliography{anthology,custom}
\bibliographystyle{acl_natbib}

\clearpage
\appendix

\section{Statistics of Bimodal Pairs}
\label{sec:appendix_pretrain_dataset}
\FloatBarrier
\begin{table}[h]
 \small
\begin{center}
\begin{tabular}{lcc}
\toprule
Language & \multicolumn{1}{l}{\# Code-Doc Pairs} & \multicolumn{1}{l}{\# Code-Comment Pairs} \\ 
\midrule
Ruby           & 48,527                            & 172,385                               \\
JavaScript     & 123,858                           & 604,678                               \\
Go             & 315,921                           & 172,385                               \\
Python         & 449,216                           & 441,976                               \\
Java           & 452,847                           & 404,424                               \\
PHP            & 520,088                           & 301,708                               \\
Overall        & 2,137,293                         & 1,964,627                             \\ \bottomrule
\end{tabular}
\caption{The statistics of code-text pairs in CodeRetriever.}
\label{table_bimodal_statistics}
\vspace{-20pt}

\end{center}

\end{table}

\section{Code-Code Pairs Construction}
\label{sec:appendix_code_code_pairs_algorithm}
\begin{algorithm}[h]
\small
\caption{\small Construct code-code pairs}
\label{pipeline_for_positive_code_pair}
\KwData{Paired text-code $(d_1,c_1),(d_2,c_2)\cdots,(d_m,c_m)$; Unpaired code data $c^*_1,c^*_2\cdots,c^*_n$.}
\KwResult{$CodePair$}
DocMatcher $\leftarrow$ SimCSE($d_1\cdots,d_m$)\;
NameMatcher $\leftarrow$ SimCSE($name_1\cdots,name_n$)\;
$CodePair_{doc}$ $\leftarrow$ []\;
$CodePair_{name}$ $\leftarrow$ []\;
\For {i $\leftarrow$ 1$\cdots$ m}{
    \For {j $\leftarrow$ i$\cdots$ m}{
        \If{sim($d_i,d_j$,$\mathrm{DocMatcher}$)>$\tau_1$}{
            $CodePair_{doc}$.append(($c_i,c_j$))
        }
    }
}

\For {i $\leftarrow$ 1$\cdots$ n}{
    \For {j $\leftarrow$ i$\cdots$ n}{
        \If{sim($name_i,name_j$,$\mathrm{NameMatcher}$)>$\tau_1$}{
            $CodePair_{name}$.append(($c^*_i,c^*_j$))
        }
    }
}

$Filter$ $\leftarrow$ CrossModel($CodePair_{doc}$)

$CodePair$ $\leftarrow$ []\;

\For{$c_i,c_j$ $\in$ $CodePair_{doc}$}{
    \If{
        Filter($c_i,c_j$) > $\tau_2$
    }{
        $CodePair$.append(($c_i,c_j$))
    }
}

\For{$c^*_i,c^*_j$ $\in$ $CodePair_{name}$}{
    \If{
        Filter($c^*_i,c^*_j$) > $\tau_2$
    }{
        $CodePair$.append(($c^*_i,c^*_j$))
    }
}

\end{algorithm}
\vspace{80pt}

\section{The Hyper-parameters for building code-code pairs.}
\label{sec:appendix_building_code_code_hyper}
\begin{table}[h]
\small
\FloatBarrier
\begin{tabular}{@{}lcc@{}}
\toprule
Hyper-parameters & Matcher & CrossModel \\ \midrule
Initialization    & GraphCodeBERT  &GraphCodeBERT  \\
Epoch             & 2            & 2    \\
Batch             & 256         & 256     \\
Learning Rate     & 2e-5        & 2e-5     \\
Optimizer & AdamW & AdamW \\
Temperature       & 0.05   & -          \\ 
Positive Threshold &0.75 &0.998\\
\bottomrule
\end{tabular}
\caption{The hyper-parameters of Matchers and CrossModel.}
\end{table}
\FloatBarrier
\section{Statistics of Unimodal Pairs}
\label{sec:appendix_unimodal_pairs}
\begin{table}[h]
\small
\setlength{\tabcolsep}{0.3mm}
\begin{tabular}{@{}lcccccc@{}}
\toprule
Language           & Ruby & JavaScript & Go & Python & Java & PHP \\ \midrule

Ruby       & 354K   & 76K         & 38K & 58K     & 78K   & 54K  \\
JavaScript & 239K   & 1936K         & 132K & 158K     & 203K   & 155K  \\
Go         & 181K   & 302K         & 3494K & 146K     & 264K   & 123K  \\
Python     & 512K   &645K         & 305K & 2038K     & 395K   & 316K  \\
Java       & 380K   & 676K         & 445K & 310K     & 4700K   & 388K  \\
PHP        & 381K   & 575K         & 241K & 246K     & 375K   & 2510K  \\
\bottomrule
\end{tabular}
\caption{The statistics of code-code pairs in CodeRetriever.}
\label{table_unimodal_statistics}
\end{table}

\section{Implementation Details}
\label{sec:appendix_implementation_details}
CodeRetriever is a siamese-encoder model with shared code encoder and text encoder. CodeRetriever is initialized with pre-trained \mbox{\textbf{GraphCodeBERT}} checkpoint released by~\citet{graphcodebert}, which is a 12 layers Transformer encoder, with hidden sizes of 768 and attention heads of 12. To save the number of model parameters, the text encoder and code encoder in CodeRetriever share their model weights during training which follows previous work~\citep{codebert,graphcodebert}.   
We use FAISS~\citep{faiss} for efficient dense indexing/retrieval. i.e., accelerate the matching of similar function names and documentations. For NameMatcher, we normalize function names according to the naming patterns. %
For example, ``openFile'' with Camel-case and ``open\_file'' with Snake-case are both normalized to ``open file''. The overall training corpus for CodeRetreiver contains 2.1 million code-doc pairs, 23.4 million code-code pairs, and 1.9 million code-comment pairs. When a code has multiple positive text or code samples, we randomly sample one of them everytime during training.
The CodeRetriever is trained with 8 NVIDIA Tesla V100s-32GB for 1.8 days.
The batch-size, learning rate and training step are 256, 4e-5 and 100K, respectively. The max sequence length of the text and code is set as 128 and 320, respectively.

\section{Statistics of Fine-tuning Data}
\label{sec:appendix_finetune_dataset}
\begin{table}[h]
 \small
\setlength{\tabcolsep}{0.5mm}
\begin{tabular}{@{}lccc@{}}
\toprule
Dataset               & Train & Dev   & Test  \\ \midrule
CodeSearch-Ruby~\citep{codesearchnet}       & 25K    & 1.4K  & 1.2K  \\
CodeSearch-JS~\citep{codesearchnet}\hspace{-10pt} & 58K    & 3.9K  & 3.3K  \\
CodeSearch-Go~\citep{codesearchnet}         & 16.7K   & 7.3K  & 8.1K  \\
CodeSearch-Python~\citep{codesearchnet}     & 25K   & 13.9K & 14.9K \\
CodeSearch-Java~\citep{codesearchnet}       & 16.4K   & 5.2K  & 10.9K \\
CodeSearch-PHP~\citep{codesearchnet}        & 24.1K   & 13.0K & 14.0K \\
Adv~\citep{codexglue}                   & 28.0K   & 9.6K  & 19.2K \\
CoSQA~\citep{cosqa}                 & 19K    & 0.5K   & 0.5K   \\
CoNaLa~\citep{conala}                & 2.8K     & -     & 0.8K   \\
SO-DS~\citep{sods}                 & 14.2K    & 0.9K   & 1.1K  \\
StaQC~\citep{staqc}                 & 20.4K   & 2.6K  & 2.7K  \\ \bottomrule
\end{tabular}
\caption{The statistics of downstream datasets.}

\end{table}

\section{Hyper-parameters of AR2}
\label{sec:appendix_ar2_hyper}
\begin{table}[h]
\small
\label{tab:ar2_hyp}
\begin{tabular}{@{}lcc@{}}
\toprule
Hyper-Parameters            & G             & D             \\ \midrule
Initialization              & GraphCodeBERT & GraphCodeBERT \\
Optimizer                   & AdamW         & AdamW         \\
Scheduler                   & Linear        & Linear        \\
Warmup proportion           & 0.1           & 0.1           \\
Negative size               & 7             & 7             \\
Batch size                  & 128           & 128           \\
Learning rate               & 5e-6          & 1e-6          \\
Max step                    & 16000         & 4000          \\ \bottomrule
\end{tabular}
\caption{The Hyper-parameters of AR2}
\end{table}

\end{document}